\documentclass[letterpaper, 10 pt, conference]{ieeeconf}  % Comment this line out if you need a4paper

\IEEEoverridecommandlockouts                              % This command is only needed if 
\usepackage{lipsum} 
\usepackage{epsfig}
\usepackage{float}
\usepackage{bbold}
\usepackage{bm}

\usepackage{color,soul} % added for yellow highlighting

\usepackage{amsmath}

\usepackage{hyperref}
\hypersetup{
    colorlinks=true,
    linkcolor=blue,
    filecolor=magenta,      
    urlcolor=cyan,
    pdftitle={Overleaf Example},
    pdfpagemode=FullScreen,
    }

\title{\LARGE \bf
Toward a Plug-and-Play Vision-Based Grasping Module for Robotics
}
\begin{document}

\author{François Hélénon$^{*\dagger1}$, Johann Huber$^{*1}$, Faïz Ben Amar$^{1}$ and Stéphane Doncieux$^{1}$ 
\thanks{$^*$ equal contribution}
\thanks{$^\dagger$ corresponding author}   \thanks{$^{1}$Sorbonne Université, CNRS, Institut des Systèmes Intelligents et de Robotique, ISIR, F-75005 Paris, France {\tt\small \{helenon, huber, benamar, doncieux\}@isir.upmc.fr}}}

\maketitle
\thispagestyle{empty}
\pagestyle{empty}

%\renewcommand{\thefootnote}{\fnsymbol{footnote}}
%\footnote[1]{text}

%%%%%%%%%%%%%%%%%%%%%%%%%%%%%%%%%%%%%%%%%%%%%%%%%%%%%%%%%%%%%%%%%%%%%%%%%%%%%%%%

%%%%%%%%%%%%%%%%%%%%%%%%%%%%%%%%%%%%%%%%%%%%%%%%%%%%%%%%%%%%%%%%%%%%%%%%%%%%%%%%
%                                 Paper's body
%%%%%%%%%%%%%%%%%%%%%%%%%%%%%%%%%%%%%%%%%%%%%%%%%%%%%%%%%%%%%%%%%%%%%%%%%%%%%%%%

%%%%%%%%%%%%%%%%%%%%%%%%%%%%%%%%%%%%%%%%%%%%%%%%%%%%%%%%%%%%%%%%%%%%%%%
%                            Abstract
%%%%%%%%%%%%%%%%%%%%%%%%%%%%%%%%%%%%%%%%%%%%%%%%%%%%%%%%%%%%%%%%%%%%%%%

\begin{abstract}

Despite recent advancements in AI for robotics, grasping remains a partially solved challenge, hindered by the lack of benchmarks and reproducibility constraints. This paper introduces a vision-based grasping framework that can easily be transferred across multiple manipulators. Leveraging Quality-Diversity (QD) algorithms, the framework generates diverse repertoires of open-loop grasping trajectories, enhancing adaptability while maintaining a diversity of grasps. This framework addresses two main issues: the lack of an off-the-shelf vision module for detecting object pose and the generalization of QD trajectories to the whole robot operational space. The proposed solution combines multiple vision modules for 6DoF object detection and tracking while rigidly transforming QD-generated trajectories into the object frame. Experiments on a Franka Research 3 arm and a UR5 arm with a SIH Schunk hand demonstrate comparable performance when the real scene aligns with the simulation used for grasp generation. This work represents a significant stride toward building a reliable vision-based grasping module transferable to new platforms, while being adaptable to diverse scenarios without further training iterations.

\end{abstract}

%%%%%%%%%%%%%%%%%%%%%%%%%%%%%%%%%%%%%%%%%%%%%%%%%%%%%%%%%%%%%%%%%%%%%%%
%                           Introduction
%%%%%%%%%%%%%%%%%%%%%%%%%%%%%%%%%%%%%%%%%%%%%%%%%%%%%%%%%%%%%%%%%%%%%%%

\section{INTRODUCTION}
Recent advances in AI have made significant progress toward building autonomous robots to release humans from strenuous tasks. Those advances include natural language-conditioned planning \cite{liang2023codeaspolicies}, foundation architectures \cite{octo2023octo}, and efficient optimization of controllers using generative models \cite{chi2023diffusion}. This progress suggests that the research field is getting closer to making robots operate in open-ended environments. However, some basic skills are only partially solved, and deploying them on a real robot requires significant engineering efforts to make them work in a given context. Grasping is an eloquent example of such a skill. While some papers demonstrate the capability to make robots learn various manipulation skills \cite{chi2023diffusion}, no off-the-shelf modules allow to address vision-based grasping on several grippers. Furthermore, the lack of proper benchmarks in physical robotics \cite{bottarel2023graspafying} prevents the community from converging on developing such a module with the best-performing approach.

Being able to rely on a visual-based grasping module would greatly interest the robotic learning field \cite{hodson2018gripping}. Grasping robustness is a critical feature of such a module. However, grasping diversity is also essential, as being able to pick up objects in a unique and context-specific way prevents some follow-up manipulation tasks.

Standard approaches for learning to grasp heavily constrain the operational space to make the problem tractable. The main design choices include top-down movements \cite{yang2023pave}, parallel grippers \cite{fang2020graspnet}, and learning from a few demonstrations \cite{qin2022from}. While leading to convincing results in the physical world, those design choices are gripper-specific and limit the adaptation capabilities of the robot and the follow-up tasks allowed by the grasps.

\begin{figure}[t]
  \centering
\centering
  \includegraphics[width=\columnwidth]{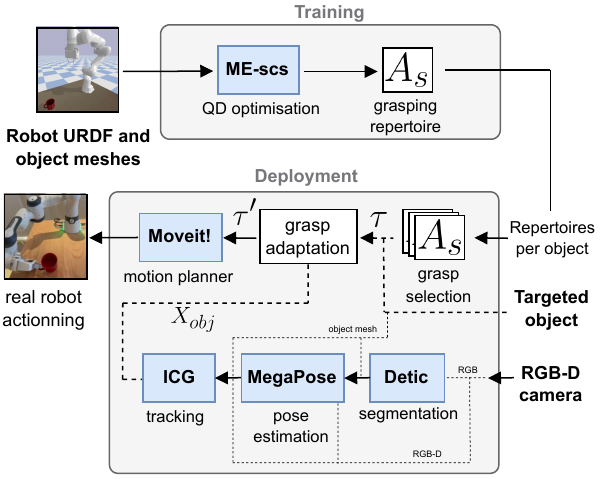}
  \caption{\textbf{Overview of the proposed framework.} It involves utilizing 3D models of the robot, target objects, and RGB-D camera data. A diverse grasping repertoire is generated with ME-scs \cite{huber2023quality} in simulation. The integration pipeline predicts the object pose through a sequence of perception modules \cite{zhou2022detic}\cite{labbe2022megapose}\cite{stoiber2022icg}. The selected grasping trajectory is transformed into the object frame and fed to a motion planner to generalize the trajectory to the whole operational space. This adaptable framework is compatible with various manipulators with minimal need for engineering efforts.}
  \label{fig:pipeline_overview}
\end{figure}

Recently, Huber et al. \cite{huber2023quality} proposed the use of Quality-Diversity (QD) methods – a family of evolutionary algorithms that aims to optimize a set of diverse and high-performing solutions to a given problem –  to generate repertoires of diverse and robust grasps for several grippers and objects. A follow-up work demonstrates the exploitability of such repertoires in the physical world through the usage of dedicated quality criteria \cite{huber2023domainrandomization}. Those methods provide an opportunity for robots to perform several grasps on a given set of objects without compromising reliability. However, a grasping repertoire generated with QD is limited to the object state (static 6D pose) initially fixed in the simulation. The computation cost of such algorithms makes it prohibitively expensive to restart the optimization process from scratch to generate grasping repertoire at any initial object state.

In this paper, we leverage QD to build diverse grasping repertoires for a set of standardized objects \cite{calli2015benchmarking} and propose an integration pipeline based on open-source state-of-the-art software to generalize the reach-and-grasps trajectories to the whole operational space of the robot. In particular:

\begin{itemize}
    \item We proposed a simple approach to generalize the reach-and-grasps trajectories to the whole operational space of the robot;
    \item We introduce an integration pipeline for exploiting those repertoires in the physical world;
    \item Experiments conducted on two robotic platforms show that this approach can efficiently be adapted to different kinds of grippers and robot arms.
\end{itemize}

The presented pipeline will be made publicly available. More details can be found on the project website\footnote{\url{https://qdgrasp.github.io/}}.

%%%%%%%%%%%%%%%%%%%%%%%%%%%%%%%%%%%%%%%%%%%%%%%%%%%%%%%%%%%%%%%%%%%%%%%
%                           Related works
%%%%%%%%%%%%%%%%%%%%%%%%%%%%%%%%%%%%%%%%%%%%%%%%%%%%%%%%%%%%%%%%%%%%%%%

\section{RELATED WORKS}
\label{sec:2_related_works}

%=====================================================================%
%                   2.1) Datasets for grasping in robotics
%=====================================================================%

%--------------------------------------------%
% Vision-based Robotic Grasping
%--------------------------------------------%
\textbf{\textit{Vision-based Robotic Grasping.}} The first paradigm for doing vision-based grasping in robotics was visual servoing based on hand-engineered features \cite{horaud1998visservoinggrasp}, but the lack of generalization made the research field move to data-driven approaches. This includes reinforcement learning \cite{chen2023rlgrasp1}\cite{zhou2023rlgrasp3}, and learning from a few demonstrations \cite{sefat2022singledemograsp}\cite{wang2021demograsp}. However, those methods are constrained by the data provided or the hand-crafted reward function, reducing the adaptation capabilities of the learned policy. To circumvent this problem, more and more works rely on data-greedy methods \cite{mousavian20196dofgraspnet}\cite{urain2023se3diffusionfield}\cite{barad2023graspldm}\cite{chen2024nsgf}. But those results are yet limited to a few simple objects, and are usually limited to 2-fingers grippers. Grasping for multi-fingers hands relies on methods based on an approximation of the object 3D bounding box and on a principal component analysis \cite{roa2012saisiedlr}\cite{grimm2021saisiekit}.

Recent works exploit automatically generated datasets to train grasp samplers \cite{weng2023ngdf}\cite{urain2023se3diff}. While those approaches are promising, transferring a learned policy to new scenes requires extensive datasets and further training to adapt to new scenarios. The present paper argues that a discrete framework is more likely to lead to a plug-and-play adaptable module in a shorter term. Ultimately, such a framework aims to make a grasping repertoire built on an FR3 arm working for operating on a table or a shelf without requiring new training iterations.

%-------------------------------------------------%
% Robotic grasping modules for the physical world
%-------------------------------------------------%

\textbf{\textit{Robotic grasping modules for the physical world.}} It is surprising how few works propose grasping integration modules, considering how crucial this skill is in robotics \cite{hodson2018gripping}. Recently, Bottarel et al. \cite{bottarel2023graspafying} proposed a benchmark protocol for robotic grasping algorithms, sharing the software material to reproduce the results on 3 standard methods. But their work evaluates data-driven approaches that require heavy computation to transfer to new scenes – if it is even allowed by the method – and their vision pipeline relies on printable markers that prevent deployment in open environments. Grasping modules used by successful teams in robotic competitions within domestic environments rely on basic approaches like manually created library of motions \cite{yi2023robocup}. Moreover, the technical reports that came out of such competitions \cite{eppner2016amazonpickchallenge}\cite{matamoros2019robocupathome} usually stick to high-level descriptions and lack publicly shared software. The purchasable robots sometimes provide built-in grasping modules. The PAL Robotics TIAGo robot \cite{tiagoPalwebsite} is provided with a grasping package that relies on primitive shapes to detect simple objects. A behavior tree is then deployed to align the gripper with the bounding box horizontally. Note that the official pick-and-place module requires a printable marker to detect the object \cite{wikiTiagoPickPlace}. The Mirokai robot \cite{encantedtoolWebsite} has a module for grasping specific tags that must be placed on any object the robot has to interact with physically. To our knowledge, none of the commonly used robots (e.g. PR2, FE Panda, Kuka, Universal Robots) are provided with a vision-based grasping module. This paper proposes a vision-based framework that inputs data from an RGB-D camera and object 3D models and produces deployable grasps from pre-trained repertoires. Flexible to the input manipulators, this module is meant to be used on various robots.

%--------------------------------------------%
% Quality Diversity
%--------------------------------------------%

\textbf{\textit{Quality Diversity.}} Quality-Diversity methods are algorithms that optimize a set of diverse and high-performing solutions to a given problem \cite{cully2022qd}. Recent works show that those methods can be used to generate repertoires of diverse reach-and-grasp trajectories \cite{huber2023quality} that can successfully be transferred in the physical world \cite{huber2023domainrandomization}. Such an approach would allow the generation of a diversity of trajectories that can fit a large variety of scenarios without new training iterations. However, the generated repertoires are limited to the initial condition of the simulated scene. Generating trajectories for a new object pose is compute-heavy. We circumvent this limitation by rigidly adapting the trajectory to the object frame using a vision-based 6 Degrees-of-Freedom (6DoF – i.e. position and orientation) object detection pipeline.

\begin{figure}[t]
  \centering
\centering
  \includegraphics[width=0.78\columnwidth]{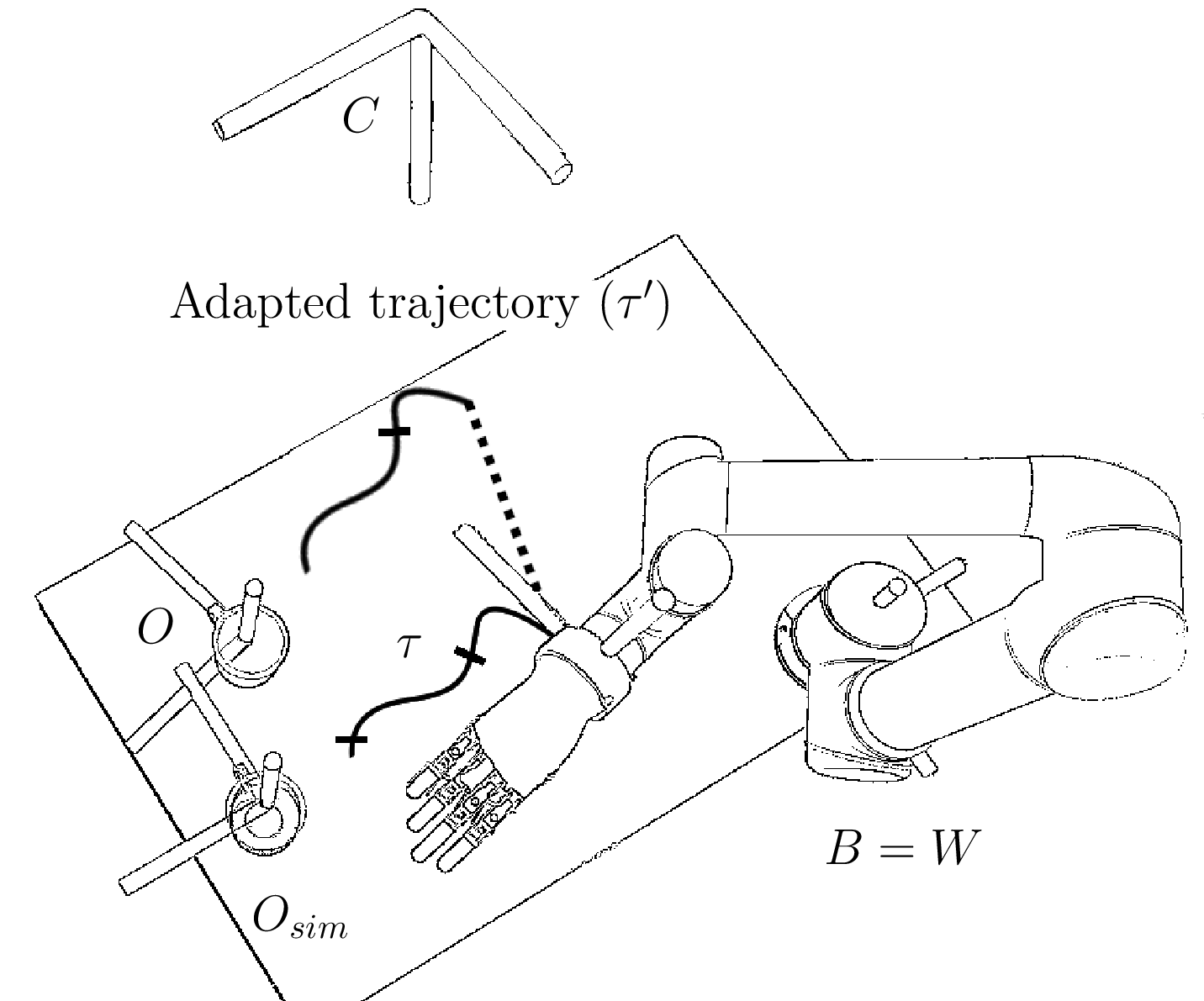}
  \caption{\textbf{Notations and adaptation principle.} The robot base frame $B$ and the world frame $W$ are assumed to be equal. The robot has to grasp a mug (frame $O$) with a pose estimated by an RGB-D camera (frame $C$) and perception modules. The trajectory $\tau$ has been generated in simulation with the object at $O_{sim}$. The path followed by the end-effector is adapted from one pose to another, resulting in the trajectory $\tau'$.}
  \label{fig:problem_definition}
\end{figure}

% À modifier sur le schéma :
% --------------------------
% Remplacer O_ref et O_target par O^sim et O
% Retirer E
% ajouter tau et tau' pour la traj initiale et la traj adaptée

\begin{figure}[t]
  \centering
\centering
  \includegraphics[width=\columnwidth]{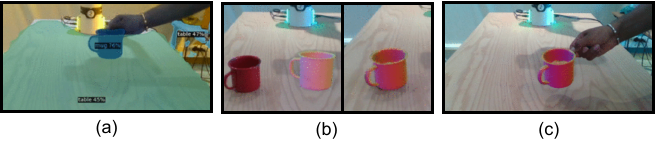}
  \caption{\textbf{Object 6DoF pose detection pipeline.} (a) The scene is first segmented to isolate the targeted object using Detic \cite{zhou2022detic}; (b) Megapose \cite{labbe2022megapose} does a 3d model matching to predict the 6DoF pose; (c) ICG \cite{stoiber2022icg} tracks the object pose to generalize the 6DoF tracking to any pose and allow retrial after failure.}
  \label{fig:object_pose_detecton_view}
\end{figure}

%%%%%%%%%%%%%%%%%%%%%%%%%%%%%%%%%%%%%%%%%%%%%%%%%%%%%%%%%%%%%%%%%%%%%%%
%                               Method
%%%%%%%%%%%%%%%%%%%%%%%%%%%%%%%%%%%%%%%%%%%%%%%%%%%%%%%%%%%%%%%%%%%%%%%

\section{METHOD}

Fig. \ref{fig:pipeline_overview} gives an overview of the proposed framework. It consists of a training step in simulation and a deployment step in the physical world. In the training step, a QD optimization method generates a set of grasping trajectories \cite{huber2023quality}. The deployment phase raises two key challenges: no off-the-shelf algorithms can robustly do the 6DoF object pose estimation, and the reach-and-grasps trajectories generated with QD are limited to a fixed initial object pose.

Fig. \ref{fig:problem_definition} shows the used notations. Let $B$ be the robot frame and $W$ be the world frame. Here, we assume that  $B=W$, as the robots considered in the experiments are fixed manipulators. Let $O_{sim}$ be the object frame at initial conditions in the deterministic simulation, and $O$ be the actual frame associated with the targeted object in the physical world (indifferently noted $X_{obj}$).

\subsection{Training}

The training part is based on previous works in grasping with QD methods: the input is the 3D model of the considered robotic manipulator, as well as the 3D models of the targeted objects. Both are included in a simulated scene, on which a QD optimization method is applied to generate a repertoire of diverse and robust grasping trajectories $A_s$ \cite{huber2023quality}. The most promising trajectories can be selected among the thousands of generated ones using dedicated quality criteria \cite{huber2023domainrandomization}. The output is, therefore, a set of repertoires containing hundreds of grasping trajectories for each of the targeted objects.

\subsection{Deployment}

The deployment part of the grasping module takes as input the data from an RGB-D camera, the name of a targeted object, and the skill repertoires containing the 3D model of the objects and the grasps.

\textit{\textbf{6DoF pose estimation.}} The estimation of the object 6DoF pose $X_{obj}$ is conducted in 3 steps (Fig. \ref{fig:object_pose_detecton_view}). The targeted object is first localized on the RGB image using an open-vocabulary segmentation module \cite{zhou2022detic}. The identified region is used to restrict the search space of a 6DoF pose estimation module called MegaPose \cite{labbe2022megapose}. MegaPose matches the object 3D model projection in the image with the current data acquired through the RGB-D camera. As soon as the predicted image converges, a low processing tracking module \cite{stoiber2022icg} is used to follow the object trajectory in the RGB-D image, generalizing the detection to the whole robot field of view and allowing retrial after failure. This results in an accurate and fast estimation of the object's 6DoF pose.

\textit{\textbf{Trajectory adaptation.}} A grasping trajectory $\tau$ is then selected with respect to the addressed scenario (e.g., higher robustness, grasping an object-specific part). In the experiments were exploited the best-performing grasps w.r.t. the fitness criterion proposed in \cite{huber2023quality}.

Let $\tau\in\mathbb{R}^{m \times n}$ be a trajectory, where $m$ is the number of values to express a state pose, and $n$ is the number of considered time steps. State-of-the-art QD methods generate open-loop trajectories conditioned on a specific object pose ($O_{sim}$). The trajectory is expressed as a succession of end-effector Cartesian positions and Euler orientations ($m=6$). A QD method thus generates a set of trajectories $A_s= \left\{ \tau_{i\in\mathbb{N}^{+*}} \right\}$. Each trajectory can be expressed as a sequence of end effector state through forward kinematics, such that $\tau = \left\{ X_{i\in[0,...,n-1]} \right\}$.

The trajectories are projected in the object frame ($O$) to generalize the generated repertoire to the whole operational space. Each repertoire can thus be interpreted as bundles of trajectories that can be reached to grasp the object in a certain manner. Let ${}^We_s$ be the end effector state in homogeneous coordinates in $W$ for a given 6D state $X$ generated in simulation. The adapted state on the real object ${}^We_r$ is defined such that:
\begin{equation}
    \label{eqn:adpatAsumption}
    {}^{O_{sim}}e_s = {}^Oe_r
\end{equation}
The adapted state in $W$ can be computed as follows: 
\begin{align}
{}^We_r & = {}_C^{W}H \,\, {}_O^{C}H \,\, {}^Oe_r \nonumber \\
 & = {}_C^{W}H \,\, {}_O^{C}H \,\, {}^{O_{sim}}e_s \nonumber \\
{}^We_r   & = {}_W^{C}H^{-1} \,\, {}_C^{O}H^{-1} \,\, {}_W^{O_{sim}}H \,\, {}^{W}e_s
\label{eqn:trajAdaptFinal}
\end{align}
where transformation matrices ${}_a^{b}H$ is the transformation matrix from $a$ to $b$. The equation (\ref{eqn:trajAdaptFinal}) allows to compute the adapted trajectory $\tau'$ from $\tau$, considering that ${}_C^{O}H$ is obtained using the proposed vision pipeline, $ {}_W^{C}H$ comes from the camera calibration, and ${}_W^{O_{sim}}H$ is provided by the simulation. Each trajectory is then filtered using the following criterion:
\begin{equation}
     \label{eqn:filterCriterion}
    f_c(\tau') = f_{IK}(\tau') \wedge f_{collision}(\tau')
\end{equation}
where $\wedge$ is the logical $and$, $f_{IK}:\mathbb{R}^{m \times n}\rightarrow  \left\{0, 1\right\}$ assesses that the trajectory is kinematically feasible with a limited jump in joint space, and $f_{collision}:\mathbb{R}^{m \times n}\rightarrow  \left\{0, 1\right\}$ assesses that no collisions happen between the robot and the environment or itself. The resulting $\tau'$ adapted trajectory can be provided to a motion planner \cite{coleman2014moveit} to complete the grasp.

%%%%%%%%%%%%%%%%%%%%%%%%%%%%%%%%%%%%%%%%%%%%%%%%%%%%%%%%%%%%%%%%%%%%%%%
%                              Experiments
%%%%%%%%%%%%%%%%%%%%%%%%%%%%%%%%%%%%%%%%%%%%%%%%%%%%%%%%%%%%%%%%%%%%%%%

\begin{figure}[t]
  \centering
\centering
  \includegraphics[width=0.8\columnwidth]{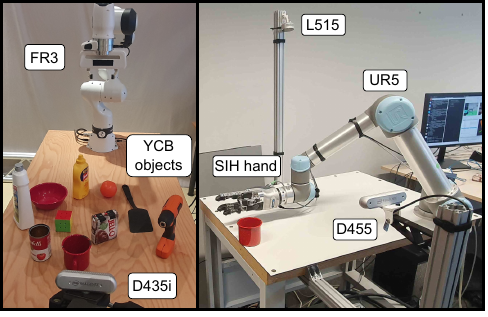}
  \caption{\textbf{Experimental setups.} To demonstrate the framework flexibility to platforms, experiments have been conducted on an FR3 arm with a parallel gripper and on a UR5 arm with an SIH 5-fingers hand. The 3 RGB-D cameras have been indifferently used to demonstrate both hardware and point-of-view robustness. The 10 YCB objects \cite{calli2015benchmarking} are used in both setups.}
  \label{fig:experimental_setups}
\end{figure}

\section{EXPERIMENTS}
\textit{\textbf{Robots and scene.}} To evaluate the proposed pipeline, experiments were carried out on a 6-DoF Universal robot (UR5) and a 7-DoF Franka Research 3 (FR3) (Fig. \ref{fig:experimental_setups}). The FR3 is equipped with a parallel 2-fingers gripper. The UR5 gripper is a SIH dexterous hand. Grasp learning and control of the SIH hand were made with synergies primitives (thumb-index, thumb-mid, thumb-index-mid, all-hand). ROS is used to orchestrate the different modules: robot and gripper control, the camera sensors, and perceptual modules. Each robot is mounted on a table, which is modeled as a collision plane.

\textit{\textbf{Sensing}} Experiments with the FR3 was conducted with a static Intel\textsuperscript \textregistered Realsense\textsuperscript T\textsuperscript M  Depth Camera D435i. For the UR5 and the dexterous hand, a Realsense D455 and a Realsense L515 were alternatively used to assess the robustness to camera point-of-view. Cameras were fixed at various mounting positions (from the top, at 45°, from the side – Fig. \ref{fig:experimental_setups}). All cameras were hand-eye calibrated with an ArUco marker.

\textit{\textbf{Hardware compute specifications}} Trajectory loading and transformations, 6-DoF tracking processing, and control of robots were made on a DELL laptop (a 12 cores Intel\textsuperscript{R} Core\textsuperscript{TM} i7-10850H). Deep learning perceptual modules were run on a remote desktop PC with a dedicated GPU (Graphical Processing Unit), an Nvidia TITAN X 12GB for the FR3, and an NVIDIA RTX 2080 for the UR5.

\textit{\textbf{Dataset generation}} Grasping repertoires were generated on the Pybullet simulator \cite{coumans2016pybullet}. ME-scs, a variant of MAP-Elites \cite{mouret2015mapelites}, was used to generate the grasps, as it appeared to be the most efficient QD method on this task \cite{huber2023quality}. The experiments were conducted on a dozen of YCB objects \cite{calli2015benchmarking}. As the YCB objects' center of mass and inertia matrix were not correctly specified in the original dataset, we computed them by getting the average position of mesh vertices and by assuming that the objects' density was $1.5kg/m^{3}$.

\textit{\textbf{Evaluating adaptation in simulation.}} To quantify the augmentation potential of the adapted trajectories, we first simulated trajectory adaptation for three objects (mug, power drill, and pudding box) in the FR3 scene. We defined a working space for the FR3 as a box in front of it. The working space was divided into equal-sized cells, defining different positions and orientations for the objects. Five trajectories were randomly sampled for each object and then adapted for each pose. For each pose, we measured the number of trajectories that were successfully adapted (i.e., the planner found a solution) – indicating the ability to generate a diversity of grasping at several positions in the working space.

\textit{\textbf{Exploitation on real platforms.}} For each object, we randomly sampled trajectories among the best-performing ones with respect to Huber et al. quality criterion \cite{huber2023quality}, promoting diversity at various object states in the working space. Each trajectory was tested with the object moved at different locations. Overall, we collected about 300 trajectories.

For pose detection, we first give the name of the object to the open-vocabulary semantic module, which then feeds the 6-DoF pose estimation pipeline with the cropped RGB-D data. Visual ambiguities can lead to the wrong initial pose estimation depending on initialization and the current view of the object. To mitigate this limitation, objects were oriented so that several faces were in the camera's field of view, and the initial pose estimation was reinitialized until convergence. Then, by leveraging the 6-DoF tracker, we placed the object at the target location. A GUI was developed to monitor the whole process (see the attached video){\footnote{\url{https://cloud.isir.upmc.fr/s/sqXpAtrrkSiM3SX}}}.

%%%%%%%%%%%%%%%%%%%%%%%%%%%%%%%%%%%%%%%%%%%%%%%%%%%%%%%%%%%%%%%%%%%%%%%
%                        Results and discussion
%%%%%%%%%%%%%%%%%%%%%%%%%%%%%%%%%%%%%%%%%%%%%%%%%%%%%%%%%%%%%%%%%%%%%%%

\begin{figure}[t]
  \centering
\centering
  \includegraphics[width=0.8\columnwidth]{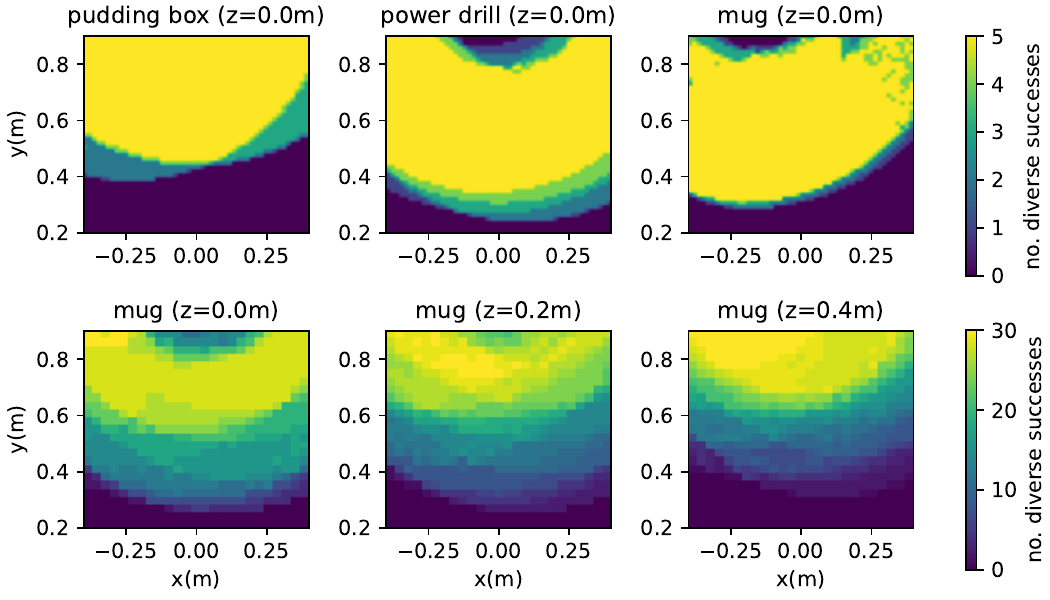}
  \caption{\textbf{Adaptation of diverse trajectories.} Results obtained in simulation on the FR3 robot by randomly picking 5 reach-and-grasp trajectories from a learned repertoire and different object poses. (Upper row): 2500 positions in the $xy$ grid at $z=0$ and for a fixed orientation. (Lower row): 625 positions and 6 orientations per position - 2 rotations around the $y$ axis and $3$ around the $z$ axis. The maximum number of transferable trajectories per pose is then 5x6=30. The rigid transform adaptation method generalizes the grasps to the whole operational space. Failures occur when rotations prevent some grasps (e.g., collisions or reachability constraints)}
  \label{fig:sim_adaptation_heatmaps}
\end{figure}

%$(x,y,z,\theta_x,\theta_y,\theta_z)\in [-0.4,0.4]\times[0.2,0.9]\times \{0\}\times\{0\}\times\{0\}$)

%$(x,y,z,\theta_x,\theta_y,\theta_z)\in [-0.4,0.4]\times[0.2,0.9]\times \{0\}\times[0,\frac{\pi}{6}]\times[0,\frac{4*\pi}{3}]$)

\begin{figure}[t]
  \centering
\centering
  \includegraphics[width=0.8\columnwidth]{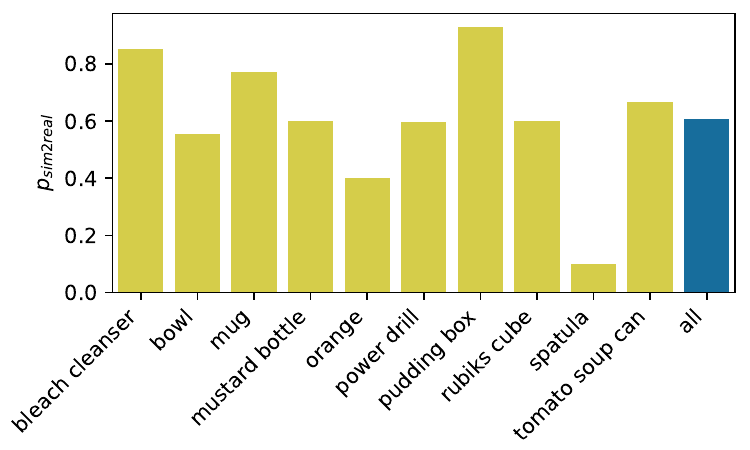}
  \caption{\textbf{Successful sim2real transfer ratios on the FR3 robot.} Results are similar to those obtained in similar experimental conditions with object fixed at simulated pose \cite{huber2023domainrandomization}, validating the proposed adaptation framework.}
  \label{fig:s2r_transfer_ratios}
\end{figure}

\begin{figure}[t]
\centering
\includegraphics[width=.5\linewidth]{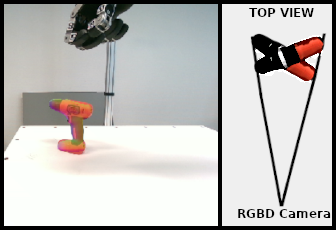}
\caption{\label{fig:challenge_vision_orientation}\textbf{Object pose ambiguities.} Ambiguities appear depending on the view, the object, and the distance to the camera. Here, the power drill is too far from the camera. Depth measurements cannot solve ambiguities. The predicted orientation is wrong over the z-axis.}
\end{figure}

\section{RESULTS \& DISCUSSION}

\textit{\textbf{Evaluating adaptation in simulation.}} Fig. \ref{fig:sim_adaptation_heatmaps} shows the results obtained in simulation for the FR3 robot with the 2-finger gripper after applying a perturbation to the object pose, either purely translated (upper row) or both translated and rotated (lower row). Most of the randomly sampled trajectories can successfully be adapted to grasp the object. The heatmap shows that the failures occur more frequently when the object is near the limits of the robot's reachable space: being too close results in self-collisions while being too far makes the robot near its joint singularities. Translating the object is likely to result in successful adaptations. It can be noticed that variations along the $z$ axis can more frequently lead to a failure. We attribute this result to singularities. Rotating the object can lead to invalid grasp because of self-occlusions (e.g. trying to grasp the hidden handle of a mug) or object poses that are outside of the collision-free and reachable space (e.g. a robot cannot grasp a cup by inserting fingers in the containing part if the object is flipped on the surface). This limitation can be addressed by regrasping or trajectory selection.

\textit{\textbf{Exploitation on real platforms.}} Fig. \ref{fig:s2r_transfer_ratios} shows the success rate of adapted grasping trajectories for all the tested setups. Most objects, even complex ones such as the power drill, show a grasp success rate of over 50\%. The average success ratio is around 60\%, which is similar to the transferability ratio obtained in the same experimental setups, except from the object pose, which matches the one in the simulation \cite{huber2023domainrandomization}. Note that the most challenging objects are the bowl, the orange, and the spatula, primarily because of vision failure. Fig. \ref{fig:challenge_vision_orientation} illustrates one example of failure modes that were observed for the power drill. As we use a single view, the partial point cloud is not always enough to disambiguate the pose. The depth measurements might also be too noisy, especially for objects small or far from the camera, making the 6-DoF pose prediction modules fall into local optima.

\begin{figure}[t]
  \centering
\centering
  \includegraphics[width=\columnwidth]{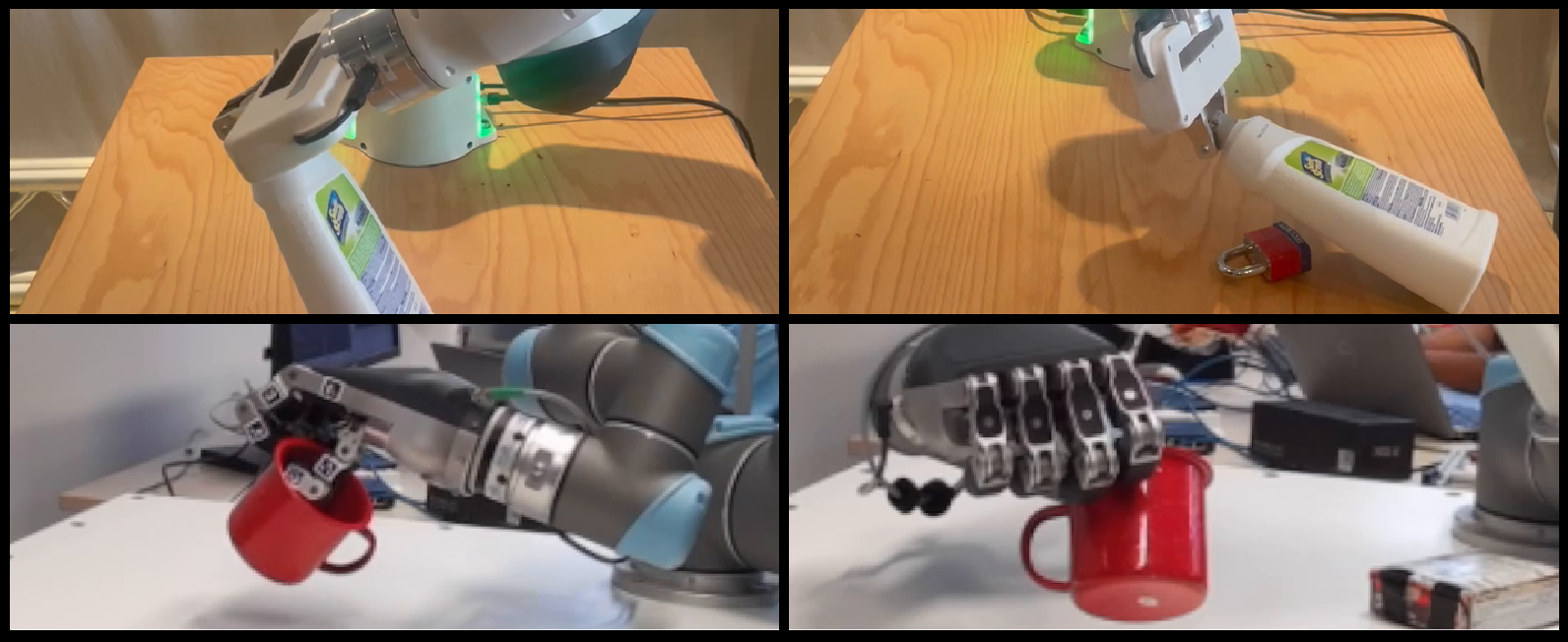}
  \caption{\label{fig:diversity_grasp_adaptation}\textbf{Examples of grasp diversity and trajectory adaptation.} The same trajectory is adapted to two different object states for the panda parallel gripper (top) or the SIH Schunk hand (bottom).}
  \label{fig:diverse_adaptation_examples}
\end{figure}

\begin{figure}[t]
  \centering
\centering
  \includegraphics[width=1\columnwidth]{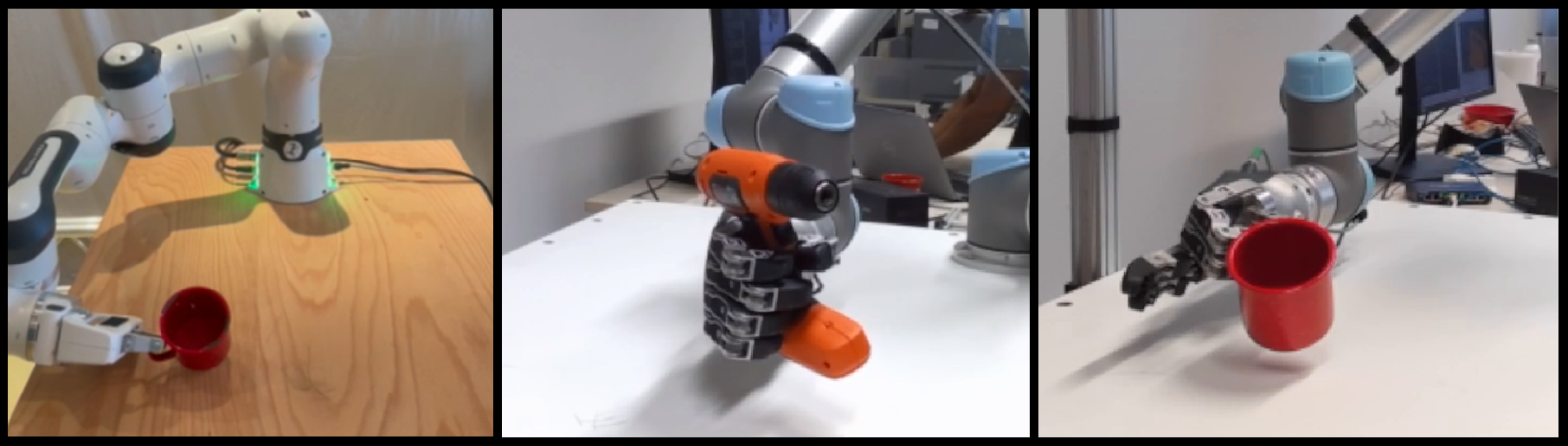}
  \caption{\textbf{Example of diverse grasps.} The diversity produced by the QD method is preserved in the proposed framework so that many different tasks can be completed after having deployed the grasping trajectory.}
  \label{fig:affordances_examples}
\end{figure}

Those experiments validate the proposed approach, as the QD-generated trajectories are well generalized to the whole operational space (Fig. \ref{fig:diverse_adaptation_examples}) with results comparable to those obtained at fixed object pose \cite{huber2023quality}. Moreover, the diversity of the generated repertoire is preserved, allowing the exploitation of the proposed framework in several scenarios without further training iterations (Fig. \ref{fig:affordances_examples}).

\textit{\textbf{Cross-plateform transferability.}} A key component of the proposed framework is how easy it is to transfer to new platforms. While the recent works on robotic learning suggest that it might be possible to exploit foundation models to do cross-platform transferability efficiently \cite{octo2023octo}, the learning methods that exploit such architectures require a tremendous amount of computation \cite{brohan2023rt2}. Moreover, it is still not clear to what extent this approach can be adapted to new scenes without a lot of computation. The approach proposed in the present paper generates the grasping repertoires offline and adapts them without additional cost. Practically, it takes a couple of hours to adapt a QD algorithm to a new robotic platform and about 5-15 minutes to generate hundreds of diverse grasping trajectories. Note that the additional integration efforts are incompressible regardless of the approach. Such flexibility can make this framework the basis of a plug-and-play vision-based grasping module.

\textit{\textbf{Limitations.}} The most important bottleneck for adaptation to a new scene is the vision pipeline, as some of the exploited submodules have limitations. In our experience, the weakest part of the vision pipeline is the 6-DoF pose estimation. MegaPose and ICG are more robust on some objects (e.g., the mug) than others (e.g., the spatula, orange). They struggle in dense or noisy scenes. It can also require a few manual iterations to converge to a valid pose. Lastly, the 3D model matching part can make prohibitive errors in the object orientations. Those errors are problematic for objects that do not have symmetrical axes (see Fig. \ref{fig:challenge_vision_orientation}). The proposed framework can, however, easily be updated with a more robust 6-DoF pose estimator and tracker. This matter is a key challenge in the computer vision community \cite{guan2024objectposevisionsurvey}. 

While easily generalizing to different robotic platforms, the proposed framework requires a 3D model of the targeted object. Nevertheless, the ability to grasp "known" objects in diverse manners opens many research paths for open-ended robotics. This limitation can be addressed by integrating vision-based surface reconstruction of unknown objects \cite{xia2024objectscan3d}.

%%%%%%%%%%%%%%%%%%%%%%%%%%%%%%%%%%%%%%%%%%%%%%%%%%%%%%%%%%%%%%%%%%%%%%%
%                             Conclusions
%%%%%%%%%%%%%%%%%%%%%%%%%%%%%%%%%%%%%%%%%%%%%%%%%%%%%%%%%%%%%%%%%%%%%%%

\section{CONCLUSIONS}

This paper proposes a framework to build a plug-and-play vision-based grasping module. It can easily be adapted to different robotic platforms and allows the robot to grasp objects in a diverse manner robustly. In future work, we plan to use dedicated quality metrics to improve the sim2real transfer ratio \cite{huber2023domainrandomization} and extend to dynamic adaptations of grasps.

%%%%%%%%%%%%%%%%%%%%%%%%%%%%%%%%%%%%%%%%%%%%%%%%%%%%%%%%%%%%%%%%%%%%%%%
%                           Acknowledgment
%%%%%%%%%%%%%%%%%%%%%%%%%%%%%%%%%%%%%%%%%%%%%%%%%%%%%%%%%%%%%%%%%%%%%%%

\section*{ACKNOWLEDGMENT}

Ministry of Education and Research (BMBF) (01IS21080), the French Agence Nationale de la Recherche (ANR) (ANR-21-FAI1-0004) (Learn2Grasp), the European Commission's Horizon Europe Framework Programme under grant No 101070381 and from the European Union's Horizon Europe Framework Programme under grant agreement No 101070596. This work used HPC resources from GENCI-IDRIS (Grant 20XX-AD011014320). Authors deeply thank Pr. Sven Behnke and the members of the AIS lab of Bonn for their warm welcome and support with the SIH setup.


%%%%%%%%%%%%%%%%%%%%%%%%%%%%%%%%%%%%%%%%%%%%%%%%%%%%%%%%%%%%%%%%%%%%%%%%%%%%%%%%

%\clearpage
%\appendices
%\input{tex_files/supp_mat/0_supp_mat_main}

\end{document}